# ICH-Qwen: A Large Language Model Towards Chinese Intangible Cultural Heritage

Wenhao Ye[1], Tiansheng Zheng[1], Yue Qi[1], Wenhua Zhao[1], Xiyu Wang[1], Xue Zhao[1], Jiacheng He[1], Yaya Zheng[2], Dongbo Wang[1*]

1. Nanjing Agricultural University, Nanjing, Jiangsu, China;
2. Nanjing Institute of Technology, Nanjing, Jiangsu, China

**Abstract:** The intangible cultural heritage (ICH) of China, a cultural asset transmitted across generations by various ethnic groups, serves as a significant testament to the evolution of human civilization and holds irreplaceable value for the preservation of historical lineage and the enhancement of cultural self-confidence. However, the rapid pace of modernization poses formidable challenges to ICH, including threats damage, disappearance and discontinuity of inheritance. China has the highest number of items on the UNESCO Intangible Cultural Heritage List, which is indicative of the nation's abundant cultural resources and emphasises the pressing need for ICH preservation. In recent years, the rapid advancements in large language modelling have provided a novel technological approach for the preservation and dissemination of ICH. This study utilises a substantial corpus of open-source Chinese ICH data to develop a large language model, ICH-Qwen, for the ICH domain. The model employs natural language understanding and knowledge reasoning capabilities of large language models, augmented with synthetic data and fine-tuning techniques. The experimental results demonstrate the efficacy of ICH-Qwen in executing tasks specific to the ICH domain. It is anticipated that the model will provide intelligent solutions for the protection, inheritance and dissemination of intangible cultural heritage, as well as new theoretical and practical references for the sustainable development of intangible cultural heritage. Furthermore, it is expected that the study will open up new paths for digital humanities research.

**Keywords**：Intangible Cultural Heritage, Large Language Modelling, Intelligent Conservation, Digital Humanities, ICH-Qwen

## 1. INTRODUCTION

Intangible Cultural Heritage (ICH) comprises diverse traditional cultural expressions inherited through generations by various ethnic groups, recognized as integral components of their cultural legacy, along with related tangible objects and spaces. It serves as an invaluable record of human civilizational progress. The United Nations Educational, Scientific and Cultural Organization (UNESCO) has promulgated multiple frameworks to advance ICH safeguarding, including the "Proclamation of Masterpieces of the Oral and Intangible Heritage of Humanity" (1998) and the "Convention for the Safeguarding of Intangible Cultural Heritage". By 2024, UNESCO's ICH Lists and Register encompass 788 entries, with China maintaining the world's largest inventory. As a critical constituent of China's outstanding traditional culture, the effective preservation, inheritance, and utilization of ICH hold profound significance for preserving historical continuity, reinforcing cultural confidence, and fostering intercultural dialogue.

However, ICH faces severe threats of deterioration, disappearance, and destruction,

---
[*] Db.wang@njau.edu.cn

necessitating urgent attention to its conservation and creative revitalization. In the context of rapid artificial intelligence advancement, modern technologies—particularly large language models (LLMs)—have infiltrated various domains, leading to the development and deployment of numerous domain-specific LLMs. To address the challenges and requirements in ICH preservation, this study leverages extensive open-source Chinese ICH data and harnesses the robust natural language processing, generation, and knowledge reasoning capabilities of LLMs for systematic knowledge organization. Furthermore, through the construction of instruction-tuning datasets incorporating synthetic data, we employ continued pre-training and fine-tuning techniques to develop a domain-specific LLM capable of deep semantic understanding and generating semantically coherent ICH-related knowledge. This approach provides an intelligent dissemination pathway for ICH transmission and preservation while offering novel methodological insights and practical references for digital humanities research.

## 2. RELATED WORK

### 2.1 Large Language Models

#### 2.1.1 Foundation Large Language Models

Bengio proposed a neural probabilistic language model that simultaneously learn the word feature vectors and the parameters of the probability function. [1] Compared to traditional N-gram language models, the proposed model demonstrated significantly reduced perplexity on test sets, indicating its superior ability to leverage contextual information and generalize to unseen word sequences during training. This marked the advent of applying neural network models to natural language understanding.

With the advancement of deep learning technologies, Recurrent Neural Networks (RNNs) gained prominence, and their variant, Long Short-Term Memory networks (LSTM), exhibited substantial advantages in handling contextual information [2][3]. In 2017, the emergence of the attention mechanism [4] and the Transformer architecture [5] accelerated progress rapid progress in fields such as machine translation and text generation.

In recent years, driven by enhanced computational power and the accumulation of massive datasets, large language models (LLMs) towards natural language generation have become the mainstream in natural language processing.The GPT series of models emerged as generative pre-trained model based on the Transformer decoder architecture. The release of ChatGPT marked a breakthrough in artificial intelligence, due to its robust language generation capabilities and flexible interactive mechanisms, rapidly gaining widespread user attention. Companies such as Google and Baidu subsequently launched comparable products. In 2023, OpenAI further introduced GPT-4, which significantly improved multimodal capabilities compared to GPT-3.5, heralding a new era in NLP development [6][7].

Currently, LLMs primarily adopt three architectural paradigms:

**Encoder-Decoder Architecture**: (also known as sequence-to-sequence architecture), which integrates both encoder and decoder components. The encoder processes input sequences into fixed-size representations, while the decoder generates output sequences based on these representations. The T5 model is a notable example of this architecture [8].

**Decoder-Only Architecture**: which exclusively employs the decoder component. This architecture is suited for sequence generation tasks, producing corresponding sequences from encoded inputs, with the GPT series being representative models.

**Encoder-Only Architecture:** which focuses on encoding input text via neural networks to extract semantic information, subsequently transmitting the encoded results to downstream processing modules. The BERT model exemplifies this paradigm [9].

2.2.2 Domain-Specific Large Language Models

Domain-specific large language models refer to large-scale language models that are tuned for specialized fields or industries. Compared to foundational LLMs, domain-specific LLMs exhibit superior domain expertise and practical applicability by focusing on domain-specific knowledge and skills. Garcia et al. [10] investigated the phenomenon of "latent domain-related trajectories" in LLMs, whether LLMs can inherently distinguish between queries from different domains and how these models adapt their decision-making and reasoning patterns across various domains.

Wang et al. [11] developed GujiBERT and GujiGPT series foundational models tailored for information processing of classical Chinese texts. These models accommodate both simplified and traditional Chinese characters, effectively addressing diverse natural language processing tasks for classical Chinese texts, including automatic sentence segmentation, punctuation restoration, word segmentation, part-of-speech tagging, entity recognition, and automated translation. Given the substantial variations in content and linguistic styles across classical Chinese texts, Yu et al. [12] proposed a hybrid classification model (GujiBERT-GCN-LSTM) integrating GujiBERT, Graph Convolutional Networks (GCNs), and Long Short-Term Memory (LSTM) networks. This framework categorizes classical Chinese texts into historical and non-historical genres, establishing a critical foundation for subsequent translation tasks.

To address the lack of legal knowledge in LLMs, Zhou et al. [13] constructed LawGPT, a legal domain-specific LLM via supervised fine-tuning, significantly improving performance on legal tasks. Xu Y et al. [14] compiled a large-scale academic dataset, MATHUSEREVAL, and proposed ChatGLM, a mathematics-oriented LLM validated through bilingual (Chinese-English) experiments. Luo Y et al. [15] demonstrated that a medically fine-tuned LLM facilitates the discovery of novel molecules and therapeutic targets, with broad applicability in biomedicine. These examples illustrate the pivotal role of domain-specific LLMs in advancing digital humanities, education, legal systems, and medical research.

2.3 Digital Intangible Cultural Heritage Research

2.3.1 Current Applications of Digital Technologies in ICH Preservation

Digital technologies, with their robust capabilities in information processing and storage, have pioneered novel pathways for the documentation, preservation, and dissemination of Intangible Cultural Heritage (ICH). In the safeguarding and transmission of ICH, the application of digital technologies extends beyond mere recording and storage; it encompasses the precise capture of ICH content, effective transformation and input of data resources, as well as scientific classification and long-term preservation in digital formats.

In recent years, numerous scholars and practitioners have conducted in-depth explorations in this field. For instance, Liu [16] systematically elaborated on specific strategies and measures for digitally preserving ICH through three dimensions: visualization technologies, database construction, and digital preservation platform design. Additionally, Amali et al. [17] employed Geographic Information Systems (GIS) to disseminate ICH-related information to the public, enhancing awareness of tangible site distributions and intangible cultural heritage attributes. Yue et al. [18] integrated advanced technologies such as large data analytics to introduce the traditional Qing-dynasty "MaMianQun" (a type of pleated skirt) into the virtual realm. By leveraging digital techniques, their practice activated digital cultural assets and propelled historical research on Chinese civilization to deeper levels. This case not only demonstrates the immense potential of digital technologies in ICH preservation but also provides valuable references for digitizing other ICH projects. Giovannini [19], taking the "La Passione di Sordevolo" (Sordevolo Passion Play) in Italy as a case study, illustrates how 3D modeling technology can reconstruct historical performance scenes of the play and integrate them into museum exhibitions, thereby enhancing audience engagement and emotional connection.

2.3.2 Challenges of Digital Technologies in ICH Transmission and Promotion

Despite the demonstrated efficacy of digital technologies in facilitating the transmission and promotion of ICH, significant constraints and limitations persisted prior to the advent of the large language model era.

On one hand, while digital technologies such as high-definition videography, 3D scanning, and digital modeling have partially achieved the documentation and preservation of ICH, their penetration remains relatively low. In remote regions or ICH projects with limited resources, the application of digital technologies is particularly restricted. Many ICH practitioners are elderly and face challenges in adapting to modern digital tools, posing challenges in safeguarding and transmitting these cultural practices [16].

On the other hand, even in digitized ICH initiatives, persistent challenges impede the comprehensive utilization and dissemination of digital outputs. These include rapid technological obsolescence, challenges in data storage and management, and uncertainties in ensuring long-term accessibility of digital archives [20]. Furthermore, emerging technologies such as large data and artificial intelligence, while offering novel possibilities for in-depth exploration and intelligent transmission of ICH, face practical barriers to widespread adoption due to high technical thresholds and substantial financial investments.

The evolution of LLMs and related technologies heralds transformative potential for ICH digitization. Leveraging robust natural language processing and deep learning capabilities, LLMs enable precise interpretation and digital representation of textual data encompassing oral traditions, artisanal techniques, and ritual practices. This integration not only facilitates comprehensive documentation and preservation through large-scale data processing but also transcends linguistic barriers via LLMs' multilingual proficiency, fostering global dissemination and cross-cultural dialogue. Consequently, the synergy between LLMs and ICH digitization is poised to yield unprecedented advancements in cultural preservation and transmission.

## 3. RESEARCH DESIGN

### 3.1 Research Framework

The proposed research framework aims to construct a high-quality Intangible Cultural Heritage domain-specific dataset through multi-source data acquisition, structured data cleansing, and granular annotation. Building upon this foundation, the study develops an instruction-tuned ICH-oriented LLM to enhance its capabilities in comprehension, generation, and interaction within the ICH domain, thereby providing a systematic solution for the digital preservation and innovation of ICH. The workflow is illustrated in the figure 1.

**Multi-source Data Acquisition:** A multi-source data collection strategy is implemented,

leveraging the China Intangible Cultural Heritage Network to gather multidimensional data from institutional, policy, and thematic repositories. To ensure data comprehensiveness and authority, domain-specific knowledge—such as scholarly literature on ICH—is integrated to augment the dataset's depth and expertise. Raw data undergoes standardized processing via rule-based cleansing protocols to eliminate noise, resolve inconsistencies, and enhance data quality. This step ensures the dataset's reliability for subsequent analytical and modeling tasks.

**Granular Knowledge Annotation**：During the pre-training phase, a knowledge entity annotation framework is developed for the ICH domain through quantitative analysis of curated data and consultations with domain experts. Following extended manual annotation and validation cycles, the dataset achieves fine-grained semantic labeling of ICH-related entities, establishing a robust data foundation for training the ICH-specific foundation model.

**Instruction-based Fine-tuning**：An instruction-based fine-tuning dataset is constructed, encompassing ICH-specific tasks such as knowledge-based Q&A, terminology interpretation, and scenario-driven dialogue. Through conversational data fine-tuning, the model's domain-specific comprehension, generative accuracy, and contextual adaptability are optimized. This framework not only offers a systematic approach to the digital and intelligent processing of ICH data but also pioneers innovative technological pathways for the preservation and revitalization of ICH.

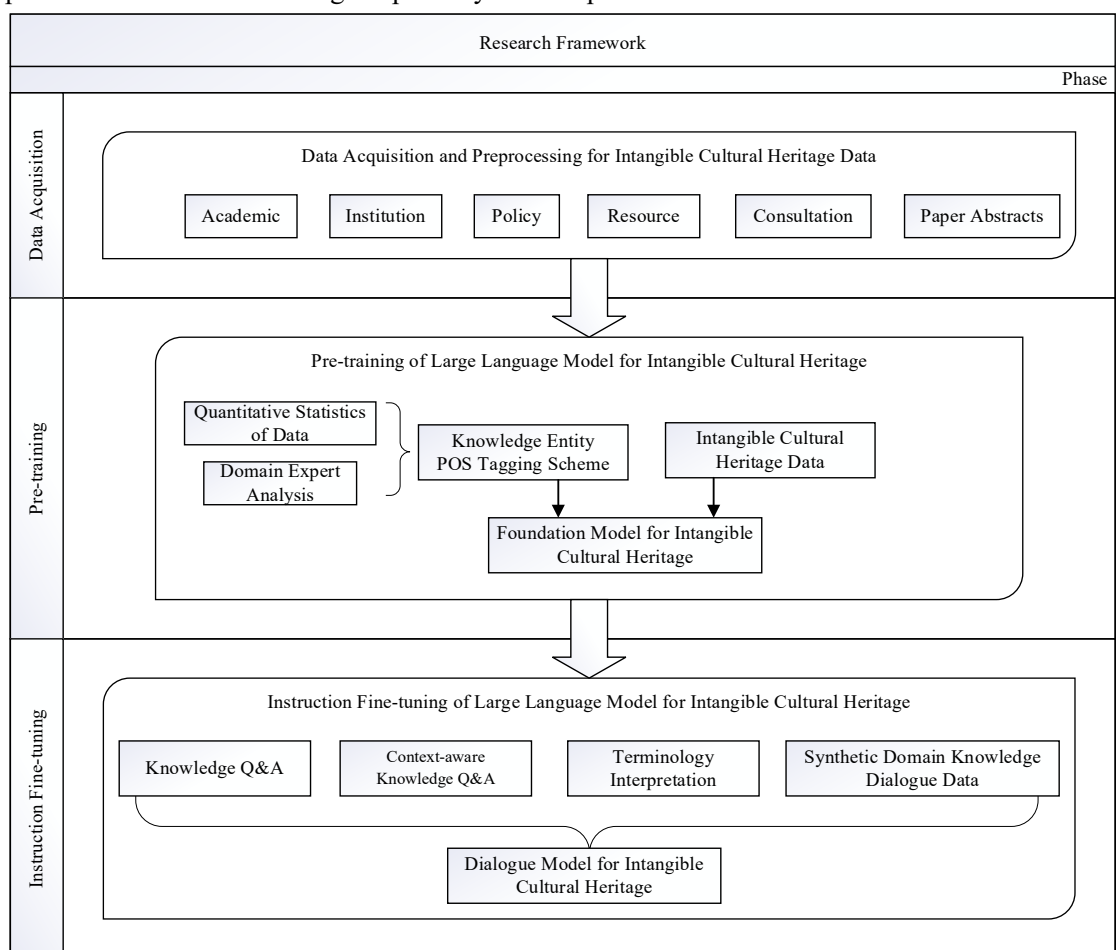

Figure1. Research Framework of ICH-Qwen Construction

3.2 Pre-training Phase

3.2.1 Data Construction

    This study employs diversified data sources, including the China Intangible Cultural Heritage

Network and academic literature databases, to conduct an in-depth analysis of the distribution, completeness, and characteristics of intangible cultural heritage data across multiple dimensions such as policy, news, and academic literature.

The China Intangible Cultural Heritage Network is recognized as one of the most authoritative sources for intangible cultural heritage data, encompassing policy regulations, news updates, special reports, and catalogs of intangible cultural heritage projects. The primary data for this study were collected from this platform, focusing on five key categories:

① Domestic Policies and Regulations: National policies, notifications, and legal frameworks related to ICH safeguarding and development.

② News Updates: Latest developments in ICH, including nationwide activities and events.

③ Thematic Reports: In-depth coverage of ICH projects, cultural preservation initiatives, and transmission practices.

④ UNESCO Documentation: Academic resources such as research papers and conference materials pertinent to ICH studies.

⑤ ICH Project Inventory: Detailed profiles of ICH projects across China, including their historical contexts, current status, and preservation strategies.

Through structural parsing of webpage content, diverse textual data were systematically extracted.

Academic literature databases provided a substantial volume of abstracts from journal articles related to intangible cultural heritage, offering critical academic support for this study. By extracting relevant journal article abstracts, the study compiled a dataset of 49,093 journal article abstracts, which comprehensively summarize the research achievements and developmental directions in the field of intangible cultural heritage. Specific statistics are presented in Table 1.

Table1. data statistics

| Data source | Categories | Num of tokens | Num of texts | Avg. length | Max length | Min length |
|---|---|---|---|---|---|---|
| The China Intangible Cultural Heritage Network | Policies and Regulations | 44945234 | 486684 | 92.35 | 386 | 54 |
| | News and Thematic Reports | 20801576 | 111782 | 186.09 | 237 | 149 |
| | Academic resources | 48767757 | 296858 | 164.28 | 497 | 25 |
| | ICH Project Inventory | 8564083 | 16695 | 512.97 | 1038 | 269 |
| Academic literature databases | abstracts from journal articles | 9604064 | 49093 | 195.63 | 312 | 146 |

### 3.2.2 Knowledge Entity Annotation

Knowledge entity annotation and part-of-speech (POS) tagging are fundamental tasks in text analysis, enabling the effective extraction of key information from large-scale textual data. For knowledge entity annotation, the study identifies and labels core entities such as ICH names, declaration locations, and ICH-specific terminology. For instance, ICH projects like "苗族古歌 (Miao Ancient Songs)" are annotated as <ICH-TITLE>. Declaration sites such as "Guizhou

Province" and "Huangping County" are labeled as <ICH-PLACE>. Terminology like "ancient songs" and "ancient lyrics" are tagged as <ICH-TERM>. This structured annotation framework facilitates precise identification of salient information in texts, supporting subsequent analytical workflows.

For POS tagging, each lexical unit is assigned a grammatical category to elucidate syntactic structures. For example, "Miao" (noun, n) denotes an ethnic group. "distributed" (verb, v) describes geographical dispersion. "widely" (adjective, a) modifies the verb "distributed". Such tagging reveals action-description relationships and entity interactions, providing a robust foundation for text comprehension, information extraction, and knowledge graph construction. The integration of knowledge entity annotation and POS tagging enables structured data extraction from ICH texts and enhances preparedness for downstream natural language processing tasks.

3.3 Instruction-based Fine-tuning

To holistically enhance the ICH-oriented foundation model's general capabilities and domain-specific knowledge, this study designs three fine-tuning tasks (see Table 2) with corresponding task-specific instructions. These instructions form the basis for constructing an instruction-based fine-tuning dataset tailored to the ICH domain. This methodological approach optimizes the model's adaptability to various application scenarios, ensuring versatility in real-world implementations.

Table 2. Fine-tuning tasks

| Evaluation Tasks | Instruction Examples | Output Examples |
| --- | --- | --- |
| Knowledge Q&A | 苗族古歌的主要演唱场合有哪些？(What are the primary performance contexts for Miao Ancient Songs?) | 苗族古歌古词神话大多在鼓社祭、婚丧活动、亲友聚会和节日等场合演唱，演唱者多为中老年人、巫师、歌手等。酒席是演唱古歌的重要场合。(Miao Ancient Songs and Myths are predominantly performed during drum society rituals, weddings, funerals, family gatherings, and festivals. Performers are typically elders, shamans, and singers. Banquets serve as significant venues for these performances.) |
| Context-aware Knowledge Q&A | <Reference Material> Using the provided content, answer the following question and ensure the output strictly derives from the given material: What are the primary performance contexts for Miao Ancient Songs? | 苗族古歌古词神话大多在鼓社祭、婚丧活动、亲友聚会和节日等场合演唱，演唱者多为中老年人、巫师、歌手等。酒席是演唱古歌的重要场合。(Miao Ancient Songs and Myths are predominantly performed during drum society rituals, weddings, funerals, family gatherings, and festivals. Performers are typically elders, shamans, and singers. Banquets serve as significant venues for these performances.) |
| Terminology Interpretation | As a professional scholar in Intangible Cultural Heritage (ICH), provide a concise introduction to the following Chinese ICH item: 苗族古歌 (Miao Ancient | 苗族古歌内容包罗万象，从宇宙的诞生、人类和物种的起源、开天辟地、初民时期的滔天洪水…今天，这些古歌古词神话还在民间流传唱诵。(Miao Ancient Songs encompass a vast array of themes, including the creation of the universe, the origins of humanity and species, mythological world-building, and ancient floods. These oral traditions continue to be recited |

| | |
|---|---|
| Songs) | and sung within Miao communities today.) |

To further enhance the domain-specific knowledge capabilities, this study constructs dialogue datasets for fine-tuning through synthetic data generation methods. Synthetic data, derived from existing datasets via data augmentation and transformation using LLMs, facilitates domain adaptation by simulating real-world conversational scenarios. During the fine-tuning process, the study leverages LLMs to synthesize three instruction-based fine-tuning datasets corresponding to the evaluation tasks of knowledge Q&A, context-aware knowledge Q&A, and terminology interpretation, thereby optimizing the model's task-specific performance. Utilizes pre-acquired ICH domain texts and archival manuscripts, this study employs Qwen2-72B-Instruct for question-answer pair generation.(see Table 3 for prompt examples).

Table 3 Prompt Examples

| | |
|---|---|
| Instruction | 我要制作一批非遗领域的知识问答数据，接下来你需要根据给出的非遗领域文本，对其进行表述和格式的修改，形成一批问答数据。<br><br>I need to generate a batch of knowledge-based Q&A data for the ICH domain. Based on the provided ICH-related texts, rephrase and reformat the content to produce standardized question-answer pairs. |

Upon inputting the ICH domain texts with the aforementioned prompts into Qwen2-72B-Instruct, the model extracts key information to construct question-answer pairs, automatically generating a substantial volume of dialogue data. Each data entry includes an input instruction and its expected output, providing explicit training objectives for model fine-tuning. Subsequently, rigorous manual verification and correction were conducted on the model-generated responses to ensure the data accurately and authentically reflect the characteristics of ICH culture, thereby meeting the high standards required for model optimization. The finalized synthetic dataset is illustrated in Table 4.

Table 4 Synthetic Data Examples

| Question | Answer |
|---|---|
| 在传统戏服中，'蟒'分为哪两种，分别适用于什么角色？ (In traditional Chinese opera costumes, into which two categories is the mang (蟒) divided, and for which roles are they respectively used?) | 在传统戏服中，'蟒'分男蟒和女蟒。男蟒是戏中帝王将相、文武百官的朝服，也是参加重大礼仪活动的礼服。女蟒则是后妃、女将等角色穿的朝服和礼服。(In traditional opera costumes, the mang is classified into male mang and female mang. The male mang serves as the court attire for emperors, generals, and officials in ceremonial contexts, while the female mang is worn by imperial consorts, female generals, and similar roles during formal occasions.) |
| 缂丝使用梭子的方法与一般织布有何不同？(How does the use of shuttles in kesi (缂丝) silk tapestry differ from conventional weaving?) | 一般织布是梭子穿过所有经线再返回，那叫通梭。但缂丝时，梭子常常是穿过一定数量的经线就返回，这叫"断纬"。'通经断纬'是缂丝特有的技法，就是经线贯穿整个布面，不同的梭子里装上所需的彩色纬线，按照不同的纹样图案，在不同的地方回返，分区分片缂织。(In conventional weaving, shuttles pass through all warp threads and return, termed tongshuo (通梭, full shuttle). In kesi, however, shuttles traverse a specific number of warp threads before returning, known as duanwei (断纬, broken weft). The distinctive 'warp-continuous, weft-broken' technique defines kesi, where warp threads span the entire fabric while colored weft threads in separate |

| 介绍一下布洛陀这一壮族神话人物。(Introduce Bulouotuo (布洛陀), a Zhuang ethnic mythological figure.) | shuttles are woven discontinuously to form intricate patterns.) 布洛陀是壮族先民口头文学中的神话人物，是创世神、始祖神和道德神。《布洛陀》是壮族的长篇诗体创世神话，主要记述布洛陀开天辟地、创造人类的丰功伟绩，自古以来以口头方式在广西壮族自治区田阳县一带传承。(Bulouotuo is a mythological deity in Zhuang oral traditions, revered as the creator deity, ancestral deity, and moral deity. The epic Bulouotuo—a lengthy poetic creation myth—chronicles his feats of cosmic creation, world-shaping, and humanity's origins. This oral tradition has been perpetuated for centuries in Tianyang County, Guangxi Zhuang Autonomous Region.) |
|---|---|

## 4. EXPERIMENTS

### 4.1 Experimental Environment

The model training and testing environment was configured as follows:

**Operating System:** Linux distribution CentOS 3.10

**GPU Specifications:** 3 × NVIDIA RTX A6000 (VRAM: 48 GB) and 6 × Quadro RTX 8000 (VRAM: 45 GB)

**Training Framework:** The LlamaFactory framework, built on the PyTorch deep learning library and integrated with the Hugging Face Transformers library to support large-scale model training and fine-tuning. The hyperparameter settings for model optimization are detailed in Table 5.

Table 5 Model Parameter Configuration

| Parameter | value |
|---|---|
| Learning Rate | 2e-4 |
| Max Epochs | 5 |
| Finetuning Type | LoRA |
| Batch Size | 4 |
| Max Sequence Length | 1024 |

### 4.2 Model Evaluation Metrics

This study utilizes several evaluation metrics to assess the model's performance across various tasks, including ROUGE-N-F, BLEU-N, and CHRF. These metrics are widely applied in natural language processing (NLP) to evaluate the quality of generated text, particularly in dialogue generation and text summarization tasks. The parameter N denotes the length of n-grams. ROUGE-N-F is a variant of the ROUGE-N metric that combines precision, recall, and F1 score to measure the coverage of key information in the generated text relative to reference texts. All metrics are grounded in n-gram analysis, with numerical ranges normalized to [0, 1]. Higher metric values indicate superior model performance. Notably, ROUGE-N-F emphasizes recall (the completeness of information), while BLEU-N focuses on precision (the accuracy of generated content).

By integrating these complementary metrics, this study provides a comprehensive assessment of the generative model's performance across diverse tasks. A series of state-of-the-art large language models, including Baichuan2-7B-Chatt, GLM-4-9B-Chat, Llama3.1-8B, Mistral-7B-Instruct-v0.3, Qwen2.5-7B-Instruct, and InternLM2.5-7B-Chat are chosed as benchmark models. These models, all released since August 2024, represent cutting-edge advancements in LLMs. Comparative analysis against these models enables a rigorous evaluation of the ICH-Qwen model's performance, leveraging their advantages in parameter scale, training data diversity, and task adaptability as robust baselines.

4.3 Pre-training Stage

As illustrated in Figure 2, the loss curve exhibits an overall downward trend, indicating the model's continuous learning and optimization. This decrease in loss reflects the model's enhanced capacity to fit the pre-training dataset specifically designed for intangible cultural heritage.

Around step 500, the loss exhibits a pronounced decline, marking a critical turning point in training. The dramatic reduction in loss suggests that the model begins to capture deeper patterns and relationships within the data. Given the complexity of intangible cultural heritage tasks, which span linguistic, historical, cultural, and sociological dimensions, this decline signifies the model's improved ability to perform semantic parsing, contextual understanding, and cultural element recognition. The convergence observed after step 800 further underscores the model's robust support for the preservation and promotion of intangible cultural heritage.

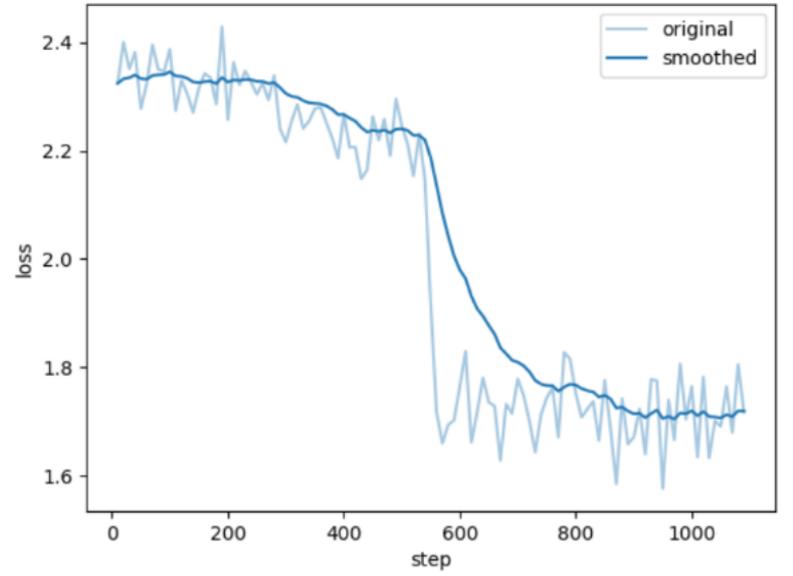

Figure 2. Training Loss Curve

4.4 Instruction Fine-tuning Stage

After pre-training, the ICH-Qwen model is fine-tuned using instruction data. To evaluate the model's performance, three task categories are defined: Knowledge Q&A, Contextual Knowledge Q&A, and Terminology Interpretation. Each task category includes 100 evaluation samples, and the performance of the ICH-Qwen model is benchmarked against state-of-the-art models, including Baichuan2-7B-Chat, Llama3.1-8B, GLM4-9B-Chat, Mistral-7B-Instruct-v0.3, Qwen2.5-7B-Instruct, and InternLM2.5-7B-Chat. This cross-model comparison enables an objective assessment of performance in the intangible cultural heritage domain. Table 6 shows evaluation results for knowledge Q&A task.

Table 6: Knowledge Q&A Evaluation Results

| Model | ROUGE-1-F | ROUGE-2-F | ROUGE-l-F | Bleu-1 | BLEU-2 | BLEU-3 | BLEU-4 | Chrf |
|---|---|---|---|---|---|---|---|---|
| ICH-Qwen | **25.04** | **7.58** | **20.88** | 18.37 | **11.91** | **8.43** | **6.32** | 10.54 |

| Model | | | | | | | |
|---|---|---|---|---|---|---|---|
| Baichuan2-7B-Chat | 23.77 | 6.07 | 16.42 | **18.44** | 11.89 | 8.15 | 5.86 | 13.43 |
| Llama3.1-8B | 9.12 | 1.96 | 8.52 | 8.11 | 5.03 | 3.45 | 2.53 | 4.24 |
| GLM4-9B-Chat | 21.17 | 5.72 | 12.31 | 14.10 | 9.39 | 6.53 | 4.80 | 12.10 |
| Mistral-7B-Instruct-v0.3 | 17.39 | 3.38 | 12.50 | 14.52 | 8.20 | 5.04 | 3.43 | 8.87 |
| Qwen2.5-7B-Instruct | 20.67 | 5.16 | 12.49 | 15.18 | 9.90 | 6.66 | 4.73 | 12.12 |
| InternLM2.5-7B-Chat | 18.70 | 4.13 | 10.23 | 12.62 | 8.02 | 5.21 | 3.55 | 10.77 |

The evaluation results demonstrate that ICH-Qwen exhibits outstanding performance across multiple metrics in Knowledge Q&A tasks. Notably, it achieves significant advantages over other models in three critical indicators: ROUGE-1-F, ROUGE-l-F and BLEU-4. This superiority indicates that the model's generated text maintains high character-level overlap with reference texts while demonstrating optimal vocabulary selection. The remarkable performance in ROUGE-l-F further reveals substantial alignment in the longest common subsequences, reflecting enhanced sentence structural similarity. The exceptional BLEU-4 score corroborates the model's advanced capability in generating long phrases and complex sentence patterns, suggesting superior matching with reference texts at character, phrasal, and syntactic levels when addressing intangible cultural heritage-related queries. Consequently, ICH-Qwen produces content that is both more accurate and linguistically fluent.

In contrast, Baichuan2-7B-Chat shows better performance in BLEU-1 and chrF metrics, slightly surpassing ICH-Qwen. This indicates its superior precision in character-level exact matching and higher similarity between generated text and reference answers. The GLM4-9B-Chat and Qwen2.5-7B-Instruct models demonstrate comparable performance, achieving moderate and relatively balanced scores across multiple metrics. Notably, the Llama3.1-8B model underperforms compared to all other models across all evaluation metrics, particularly showing substantial deficiencies in ROUGE-2-f and BLEU-1 scores. This performance gap suggests significant discrepancies in lexical selection and sentence composition between its generated text and reference standards within knowledge-intensive Q&A tasks.

Table 7: Context-aware Knowledge Q&A Evaluation Results

| Model | ROUGE-1-F | ROUGE-2-F | ROUGE-l-F | BLEU-1 | BLEU-2 | BLEU-3 | BLEU-4 | Chrf |
|---|---|---|---|---|---|---|---|---|
| ICH-Qwen | **37.21** | **10.48** | **27.74** | **29.66** | **20.82** | **15.45** | **11.99** | 16.06 |
| Baichuan2-7B-Chat | 28.12 | 8.90 | 15.49 | 12.80 | 9.97 | 7.78 | 6.17 | **16.16** |
| Llama3.1-8B | 19.33 | 7.92 | 12.83 | 10.41 | 8.29 | 6.60 | 5.36 | 10.90 |
| GLM4-9B-Chat | 19.08 | 4.36 | 8.92 | 7.48 | 5.51 | 4.00 | 2.94 | 10.26 |
| Mistral-7B-Instruct-v | 23.72 | 7.76 | 14.66 | 11.13 | 8.54 | 6.61 | 5.26 | 14.5 |

| Model | | | | | | | |
|---|---|---|---|---|---|---|---|
| 0.3 | | | | | | | 6 |
| Qwen2.5-7B-Instruct | 18.47 | 4.96 | 10.12 | 8.47 | 6.25 | 4.62 | 3.49 | 11.32 |
| InternLM2.5-7B-Chat | 16.69 | 4.17 | 8.15 | 6.75 | 5.15 | 3.82 | 2.85 | 10.06 |

As shown in table 7, in context-aware knowledge Q&A tasks, ICH-Qwen demonstrates optimal evaluation outcomes, outperforming comparative models across seven metrics. This achievement signifies its enhanced capacity to align with reference answers lexically while exhibiting robust contextual comprehension and text generation capabilities. The Baichuan2-7B-Chat model maintains its superiority in the chrF metric, consistent with its performance in standard knowledge Q&A tasks, while ranking second to ICH-Qwen in most other indicators. This pattern confirms Baichuan2-7B-Chat's persistent advantages in intangible cultural heritage domain tasks. Conversely, GLM4-9B-Chat and InternLM2.5-7B-Chat exhibit suboptimal performance across multiple metrics, particularly showing substantial performance gaps in ROUGE-2-f, BLEU-3, and BLEU-4. Such deficiencies indicate their limited capacity for generating longer n-gram sequences that align with reference texts.

Comparative analysis with standard QA tasks reveals statistically significant improvements in ICH-Qwen's performance under contextual prompting. Specifically, ROUGE-1-f increases from 25.04 to 37.21 (+48.6%) and BLEU-4 from 6.32 to 11.99 (+89.7%), indicating enhanced text generation proficiency through improved lexical matching and semantic coherence. This phenomenon may be attributed to contextual cues that provide clearer directives for generation, thereby facilitating a more precise interpretation of query intent and the subsequent production of responses that are aligned with reference standards.

Table 8: Terminology Interpretation Evaluation Results

| Model | ROUGE-1-F | ROUGE-2-F | ROUGE-l-F | BLEU-1 | BLEU-2 | BLEU-3 | BLEU-4 | Chrf |
|---|---|---|---|---|---|---|---|---|
| ICH-Qwen | **23.82** | **8.63** | **18.69** | **25.41** | **15.82** | **11.01** | **8.41** | **14.42** |
| Baichuan2-7B-Chat | 19.54 | 3.98 | 16.50 | 22.58 | 12.71 | 7.81 | 5.07 | 9.97 |
| Llama3.1-8B | 9.58 | 1.52 | 8.25 | 7.59 | 3.95 | 2.39 | 1.57 | 3.68 |
| GLM4-9B-Chat | 19.44 | 4.15 | 14.53 | 23.42 | 13.47 | 8.31 | 5.39 | 10.62 |
| Mistral-7B-Instruct-v0.3 | 16.33 | 2.49 | 13.67 | 17.80 | 8.65 | 4.76 | 2.86 | 7.40 |
| Qwen2.5-7B-Instruct | 20.20 | 4.13 | 15.10 | 23.30 | 13.13 | 7.96 | 5.07 | 10.31 |
| InternLM2.5-7B-Chat | 18.88 | 3.64 | 14.07 | 22.15 | 12.24 | 7.31 | 4.59 | 9.93 |

In the terminology interpretation task, illustrated in table 8, ICH-Qwen achieves superior performance across all metrics, demonstrating its capability to accurately comprehend terminological semantics and generate explanations that align with reference standards. The

Baichuan2-7B-Chat, Qwen2.5-7B-Instruct, and GLM4-9B-Chat models also exhibit competitive performance, though marginally inferior to ICH-Qwen across multiple metrics, suggesting room for improvement in their understanding and generation of domain-specific terminology for ICH. Notably, Llama3.1-8B underperforms significantly in all metrics, particularly showing critically low scores in ROUGE-1-f (Δ=21.3 vs. ICH-Qwen) and BLEU-1 (Δ=18.7), indicating fundamental discrepancies in character-level alignment with reference texts. Its limited capacity for longer n-gram generation further reveals deficiencies in lexical selection consistency.

ICH-Qwen demonstrates state-of-the-art performance across all three evaluation tasks. Its enhanced capability in contextualized knowledge QA tasks highlights superior adaptability to complex linguistic environments. The model's exceptional performance in terminology interpretation further validates its semantic mastery of specialized vocabulary, solidifying its position as a robust text generation system with advanced contextual understanding in ICH domains.

## 5. DISCUSSIONS

The performance of domain-specific large language models is governed by synergistic interactions among training data, base architectures, optimization strategies, and domain expertise. The ICH-oriented LLM developed in this study demonstrates effective knowledge retention and task execution capabilities, significantly advancing intelligent preservation and application of intangible cultural heritage.

Data quality constitutes the most pivotal factor for ICH-oriented LLM development. The predominantly oral transmission and multimodal nature of ICH (e.g., traditional arts, music, and craftsmanship) pose substantial data acquisition challenges. This study addresses these challenges by constructing a multidimensional ICH dataset that incorporates institutional records, policies, academic publications, and project documentation. However, the diversity of ICH manifestations—spanning textual, auditory, and visual formats across geographical regions—necessitates future integration of multimodal data (e.g., speech and video) to optimize model capabilities, a critical step toward developing multimodal ICH LLMs.

Domain-specific knowledge serves as the cornerstone for evaluating ICH model adaptation. To enhance task-specific understanding, we developed a knowledge entity annotation framework through quantitative analysis of ICH data and expert validation. This framework enabled meticulous annotation of ICH entities via iterative human-in-the-loop verification. Furthermore, leveraging generative capabilities of contemporary LLMs, we augmented training data through semantic transformations of authoritative resources (e.g., Chinese World-Class ICH Culture Series), thereby creating high-quality dialogic data for cultural dissemination. Future research will explore multilingual and dialectal adaptations to facilitate global ICH propagation.

The selection of fundation models and optimization methodologies provides technical foundations for ICH LLM deployment. Among seven evaluated dialog models, Qwen2.5-7B-Chat emerged as the optimal base architecture due to its exceptional performance in knowledge-intensive tasks. Building upon this foundation, we developed specialized ICH-Qwen models using high-performance computing infrastructure. These models establish critical benchmarks for implementing LLM technologies in ICH applications, enabling intelligent support for cultural research and public education. Subsequent work will focus on developing task-specific adaptations to deepen LLM integration within ICH ecosystems.

## 6. CONCLUSION

Globalization, while fostering modern civilization, has also precipitated a trend of global cultural homogenization, thereby posing significant threats to cultural diversity in what can be referred to as the "barbarism of modernity." Strengthening the protection of intangible cultural heritage is not only critical for the inheritance and development of Chinese culture but also plays a pivotal role in advancing the diversification of human civilization. This study addresses the specific needs of the ICH domain by constructing a robust large language model tailored for ICH. The deployment of this model can significantly advance the digital preservation of ICH, enabling systematic digitization, knowledge organization, and institutionalization of traditional stories, skills, customs, and other cultural elements. This contributes to the establishment of a sustainable protection framework for ICH, ensuring its long-term viability and development.

In the context of ICH dissemination, the ICH-Qwen model facilitates automated cultural propagation and knowledge popularization. Its advanced cross-linguistic capabilities enable the integration of regional ICH traditions, overcoming language barriers and fostering international recognition of ICH. This enhances the global outreach of ICH and supports the construction of cultural confidence for the Chinese nation.

Regarding the support of ICH inheritance, the ICH-Qwen model offers technical assistance to cultural practitioners, empowering them to utilize emerging technologies for the creative transformation of ICH. Furthermore, the study promotes cultural re-creation and facilitates an in-depth exploration of the intrinsic spiritual values embedded within ICH. Combined with knowledge graph technology, this approach facilitates systematic knowledge storage and analysis of ICH, offering researchers a structured methodology to investigate the internal logic and transmission pathways of intangible cultural heritage.

**Author contributions**

Dongbo Wang designed the overall research framework. Wenhao Ye and Tiansheng Zheng trained the ICH-Qwen model and drafted the manuscript. Yue Qi and Wenhua Zhao conducted the evaluation experiments. Xiyu Wang validated the experimental results. Xue Zhao, Jiacheng He, and Yaya Zheng revised and finalized the paper.

**Data availability**

The datasets used and analysed during the current study available from the corresponding author on reasonable request.

## ACKNOWLEDGMENTS

The research was funded by the National Social Science Foundation of China (No.21&ZD331) and the Social Science Fund Project of Jiangsu Province(No. 23TQC004).